\documentclass[letterpaper, 10 pt, conference]{ieeeconf}  

\IEEEoverridecommandlockouts                              

\input{bib_short.def}

\usepackage{graphics} 
\usepackage{epsfig} 
\usepackage{mathptmx} 
\usepackage{times} 
\usepackage{amsmath} 
\usepackage{amssymb}  
\usepackage{booktabs}
\usepackage{multirow} 
\usepackage{makecell}
\usepackage{tikz}

\usepackage{xcolor,colortbl}
\definecolor{Gray}{gray}{0.85}
\definecolor{LightCyan}{rgb}{0.88,1,1}

\newcolumntype{a}{>{\columncolor{Gray}}c}
\newcolumntype{b}{>{\columncolor{white}}c}

\newlength\barwidth  
\newlength\barheight 

\definecolor{colA}{RGB}{31,119,180}
\definecolor{colB}{RGB}{255,127,14}
\definecolor{colC}{RGB}{44,160,44}

\newcommand{\BarSparkA}[1]{%
  \begin{tikzpicture}[x=\barwidth,y=\barheight,baseline=(current bounding box.south)]
    \fill[colA] (0,0)   rectangle (1,#1/100);
    \node[below, font=\small] at (0.5,-0.05)   {#1};
  \end{tikzpicture}%
}
\newcommand{\BarSparkB}[1]{%
  \begin{tikzpicture}[x=\barwidth,y=\barheight,baseline=(current bounding box.south)]
    \fill[colB] (0,0)   rectangle (1,#1/100);
    \node[below, font=\small] at (0.5,-0.05)   {#1};
  \end{tikzpicture}%
}
\newcommand{\BarSparkC}[1]{%
  \begin{tikzpicture}[x=\barwidth,y=\barheight,baseline=(current bounding box.south)]
    \fill[colC] (0,0)   rectangle (1,#1/100);
    \node[below, font=\small] at (0.5,-0.05)   {#1};
  \end{tikzpicture}%
}

\title{\LARGE \bf
Beyond Overall Accuracy: Pose- and Occlusion-driven Fairness Analysis in Pedestrian Detection for Autonomous Driving
}

\author{Mohammad Khoshkdahan$^{1}$, Arman Akbari$^{2}$, Arash Akbari$^{2}$, Xuan Zhang$^{2}$
\thanks{$^{1}$Mohammad Khoshkdahan is with the Institute of Applied Informatics and Formal Description Methods, Karlsruhe Institute of Technology, Karlsruhe, Germany. {\tt\small mohammad.khoshkdahan@kit.edu}}%
\thanks{$^{2}$Prof. Dr. Xuan Zhang, Arash Akbari, and Arman Akbari are with the Department of Electrical and Computer Engineering, Northeastern University,
        MA, USA.
        }%
}

\begin{document}

\maketitle
\thispagestyle{empty}
\pagestyle{empty}

\begin{abstract}

Pedestrian detection plays a critical role in autonomous driving (AD), where ensuring safety and reliability is important. While many detection models aim to reduce miss-rates and handle challenges such as occlusion and long-range recognition, fairness remains an underexplored yet equally important concern. In this work, we systematically investigate how variations in the pedestrian pose—including leg status, elbow status, and body orientation—as well as individual joint occlusions, affect detection performance. We evaluate five pedestrian-specific detectors (F2DNet, MGAN, ALFNet, CSP, and Cascade R-CNN) alongside three general-purpose models (YOLOv12 variants) on the EuroCity Persons Dense Pose (ECP-DP) dataset. Fairness is quantified using the Equal Opportunity Difference (EOD) metric across various confidence thresholds. To assess statistical significance and robustness, we apply the Z-test. Our findings highlight biases against pedestrians with parallel legs, straight elbows, and lateral views. Occlusion of lower body joints has a more negative impact on the detection rate compared to the upper body and head. Cascade R-CNN achieves the lowest overall miss-rate and exhibits the smallest bias across all attributes. To the best of our knowledge, this is the first comprehensive pose- and occlusion-aware fairness evaluation in pedestrian detection for AD.

\end{abstract}

\section{INTRODUCTION}
Pedestrian detection is a critical component of autonomous vehicles (AVs), which ensures safe navigation by preventing collisions in urban environments \cite{adchallenges}. ML-based detectors analyze camera images to detect pedestrians and provide confidence scores for their predictions. Researchers try to lower the overall miss-rate of the detectors, defined as the proportion of undetected pedestrians relative to all instances. However, equitable performance across subgroups is also essential for trustworthy autonomous driving (AD). Disparities in detection, such as those based on gender, can compromise pedestrian safety and reduce public trust in the technology.

Empirical studies reveal several forms of bias for age, gender, and skin color. For instance, children are missed more often than adults \cite{chen2024fairness}, and detectors have a higher miss-rate on pedestrians with darker skin tones \cite{wilson2019predictive}. Such disparities must be addressed to achieve equitable safety. Besides, occlusion poses another challenge. While many methods achieve strong overall accuracy under partial occlusion, prior works have not systematically explored how occluding individual joints (e.g., hips or ankles) might impact detection performance.

\begin{figure}[t]
    \centering
    \includegraphics[width=1\linewidth]{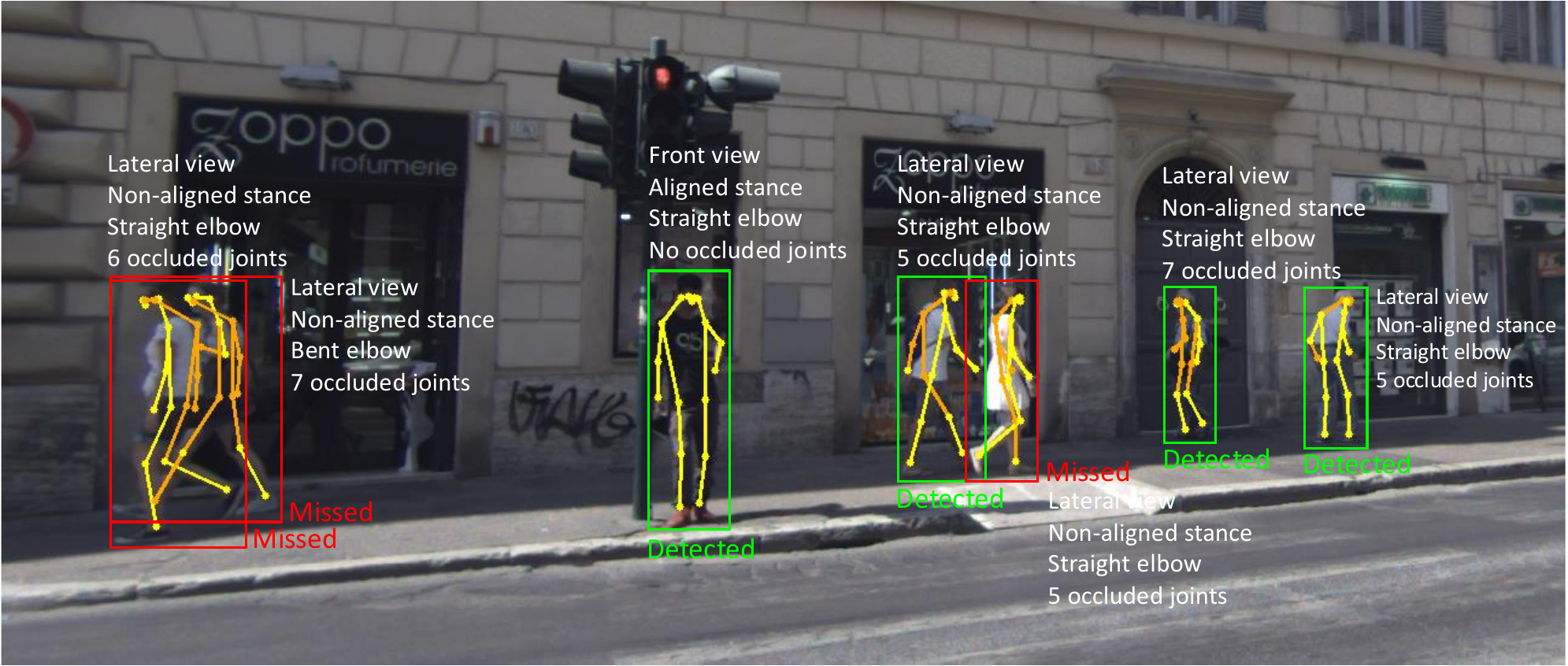}
    \caption{Sample scene image from the ECP-DP \cite{braun21simple}, annotated with joints. Yellow joints are visible, and orange ones are occluded. Green boxes show correct detections, and red ones show missed pedestrians (by YOLOv12-S \cite{yolov12}). Automatically extracted pose-related attributes appear in white text.}
    \label{fig:motiv_img}
\end{figure}

In addition, the vehicle might perceive pedestrians with different postures (see Fig. \ref{fig:motiv_img}). For instance, walking and standing produce different leg configurations, or holding a phone or using objects can alter arm posture. These pose-related factors may introduce biases that have yet to be examined in prior works.

Sources of unfairness include both data and model design. Naturalistic data collection might not yield a balanced sample set for all sensitive attributes. For example, a system trained mostly on front-facing pedestrians might struggle with those viewed from the side (different visible silhouettes). Model's architectural choices can also introduce biases in detection.

\begin{figure*}[t]
\vspace{0.2cm}
    \centering
    \includegraphics[width=1.0\linewidth]{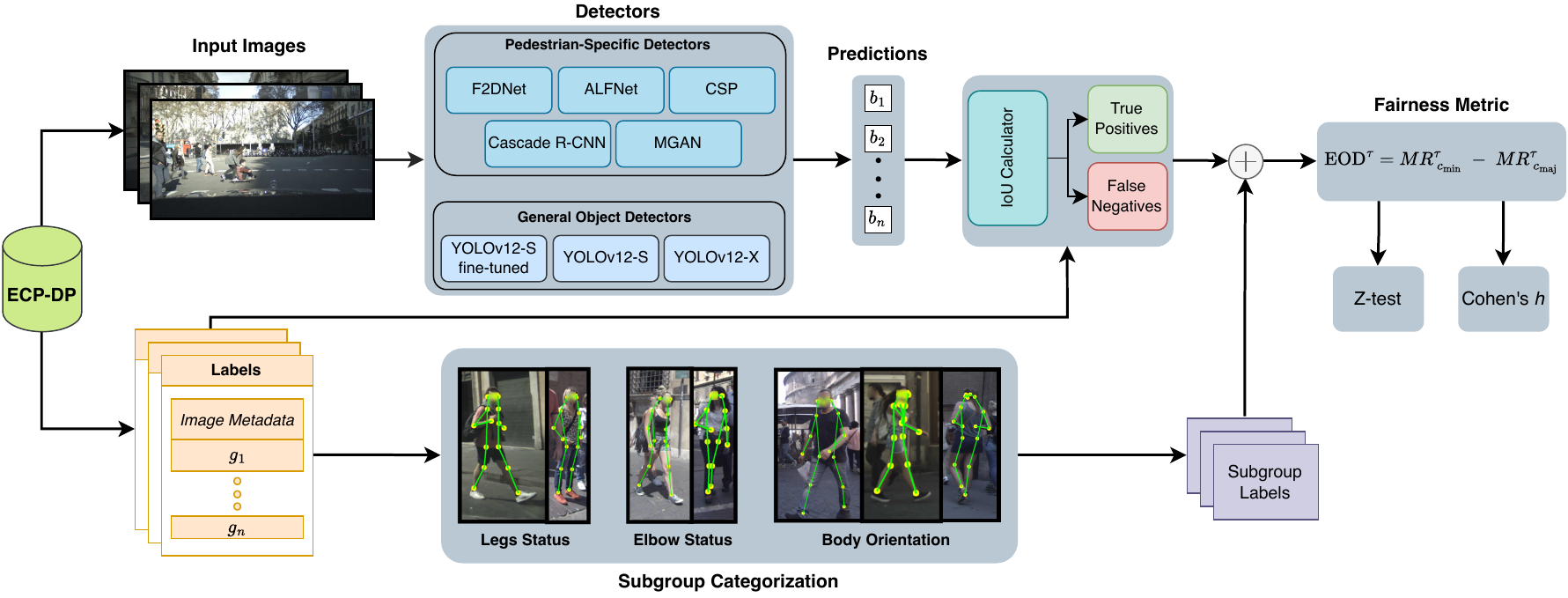}

    \caption{Pose- and Occlusion-Driven Fairness Analysis Framework. Top: ECP-DP \cite{braun21simple} images are processed by eight detectors and matched to ground truths via IoU to compute true positives and false negatives per subgroup. Bottom: Pose attributes (leg status, elbow status, body orientation) and occlusion flags for 17 joints are extracted from annotations. Subgroup detection results are then evaluated with the fairness metric to quantify bias.}

    \label{fig:overview-graph}
\end{figure*}

This work fills these gaps and contributes to fairness evaluation in pedestrian detection through three key steps. First, we evaluate eight detection methods on the ECP-DP dataset~\cite{braun21simple}, including five pedestrian-specific detectors, two state-of-the-art general-purpose detectors, and a general object detector fine-tuned on pedestrian data. Second, we automatically annotate four pose-related attributes—leg status, elbow status, body orientation, and joint occlusions—to compute subgroup miss-rates, employ fairness metrics to quantify biases across all attributes, and validate findings with statistical tests. Third, we analyze how bounding-box height (instance size) affects subgroup miss-rate disparities to determine whether detection bias persists even when controlling for pedestrian size. Our work systematically compares diverse detectors, identifies pose-related biases, and provides quantified results through the proposed pipeline (see Fig. \ref{fig:overview-graph}).

\section{RELATED WORKS}
Modern pedestrian detection systems predominantly employ deep learning architectures, particularly convolutional neural networks (CNNs) and transformer-based models \cite{modern}. These include both dedicated pedestrian detectors and general-purpose (multi-class) object detectors.

Pedestrian-specific models have been explicitly designed to tackle challenges unique to pedestrian detection. F2DNet~\cite{f2dnet} introduces a fine-to-dense representation strategy to improve detection under severe occlusion. MGAN~\cite{mgan} adopts a multi-granularity attention mechanism to enhance feature extraction across varying pedestrian scales. ALFNet~\cite{alfnet} proposes an anchor-level feature fusion framework that improves localization accuracy, particularly for small and occluded targets. CSP~\cite{csp} uses a coarse-to-fine scale perception structure to adaptively capture pedestrians at different resolutions. Pedestron~\cite{pedestron} adopts Cascade Mask R-CNN as its core detector, applying a progressive training strategy that pre-trains on diverse large-scale datasets before fine-tuning on pedestrian-specific ones like CityPersons~\cite{cityperson}.

In contrast, general-purpose detectors such as YOLOv12~\cite{yolov12}, originally developed for multi-class detection on datasets like COCO~\cite{coco}, have been adapted for pedestrian tasks through fine-tuning. YOLO~\cite{yolo} models detect objects in a single pass over the image, achieving real-time performance by directly regressing bounding boxes and class scores. Particularly, YOLOv12~\cite{yolov12} integrates an attention-centric architecture into the YOLO framework, setting a new benchmark for optimizing both inference speed and detection accuracy in real-time applications.

Fairness in machine learning faces challenges from representation, measurement, and algorithmic biases \cite{mehrabi2021survey}. Algorithmic bias emerges when model training and optimization resolutions unintentionally favor majority subgroups, while representation bias is introduced by datasets that do not accurately capture real-world demographic variance. In pedestrian detection by AVs, fairness remains understudied despite its safety-critical role. Beyond onboard perception in AVs themselves, fairness also matters when detection is done by external agents, like traffic moderator robots \cite{alexey-IAVVC25, Alexey-ELMAR25}, making it a broader and important topic. Prior works focus on demographic attributes like age, gender, and skin tone. For example, detection models show a 1.19\% higher miss-rate for female pedestrians than males \cite{li2024bias}, overlook children approximately 1.3 to 1.4 times more than adults \cite{agebias}, and achieve 2.8\% lower precision for darker-skinned individuals in low-light conditions \cite{wilson2019predictive}. A recent study \cite{li2024bias} revisits these same attributes under varying weather conditions, revealing stronger biases at night. However, this work relies on manual labeling limited to less than 5,000 pedestrian samples (per dataset) and does not include pose-related factors. To make the investigations more valid, our work discovers biases in \(N\approx 1.9\times10^5\) pedestrian samples. This enables large-scale analysis of unexplored biases, addressing prior limitations in dataset size and subgroup diversity.

Existing methods for pedestrian detection improve occlusion handling through model architecture \cite{mgan} or data augmentation \cite{augmentation}, but they focus on overall accuracy rather than fairness when specific joints are occluded. A recent study \cite{fair-ped} has proposed a novel pipeline for fairness evaluation based on visual attributes. However, no prior work has comprehensively studied how occlusion of each individual joint impacts detection performance differently. Hence, this gap is also addressed in our work through detailed analysis.

\section{METHODOLOGY}

We propose a fairness framework (see Fig. \ref{fig:overview-graph}) to evaluate pose-related biases in pedestrian detection using four critical attributes: leg status, elbow status, joint occlusion, and body orientation. All instances in the ECP-DP~\cite{braun21simple} dataset are automatically labeled with these attributes. We then assess detection performance across attribute subgroups using eight different detectors. The biases are then quantified through our fairness metric and validated by statistical tests.

\subsection{Subgroup Categorization}

\subsubsection{Legs Status}

We assign each pedestrian a binary leg label \(S\in\{\emph{Aligned stance},\emph{Non-aligned stance}\}\). We first check knee flexion. If neither knee is flexed beyond a threshold, we subsequently evaluate leg separation and line intersection.

Let \({H}_k, {K}_k, {A}_k\) denote the hip, knee, and ankle points respectively for each side \(k \in \{\text{left}, \text{right}\}\). We define the angle between two vectors \(\vec{a}\) and \(\vec{b}\) as:
\begin{equation}
\label{eq:angle_eq}
  \measuredangle(\vec{a}, \vec{b})
  = \cos^{-1}\!\biggl(\frac{\vec{a}\cdot\vec{b}}{\|\vec{a}\|\;\|\vec{b}\|}\biggr).
\end{equation}

The hip-ankle vectors are defined as \(\vec{v}_{\text{left}} = A_{\text{left}} - H_{\text{left}}\) and \(\vec{v}_{\text{right}} = A_{\text{right}} - H_{\text{right}}\). The knee angles \(\phi_{\text{left}}\) and \(\phi_{\text{right}}\) are computed as \(\phi_k = \measuredangle(K_k - H_k, A_k - K_k)\) for \(k \in \{\text{left}, \text{right}\}\). The enclosed angle between the hip-ankle vectors is \(\theta = \measuredangle(\vec{v}_{\text{left}}, \vec{v}_{\text{right}})\). This angle \(\theta\) indicates how spread the legs are in the view plane (e.g., parallel or open from the hips). We set the knee flexion threshold \(\gamma=12^\circ\) and the hip-ankle angle threshold \(\alpha=10^\circ\). We also find the intersection point \((x_I, y_I)\) by solving the two line equations for \(\vec{v}_{\text{left}}\) and \(\vec{v}_{\text{right}}\).

Consequently, we label each pedestrian’s stance as follows. A pedestrian is assigned \emph{Non-aligned stance (NAS)} if at least one knee is flexed beyond \(\gamma\), or if the hip–ankle separation angle \(\theta\) exceeds \(\alpha\) and the intersection point \((x_I,y_I)\) lies above the ankle level. Otherwise—when both knees are straight and the legs are close together (parallel or with the intersection at or below ankle level)—the pedestrian is assigned \emph{Aligned stance (AS)}.

Let $y_{\mathrm{ank}} = \min\{y_{A_L},\,y_{A_R}\}$ denote the higher ankle's vertical coordinate. We classify stances as either \emph{NAS} or \emph{AS}:

\begin{equation}
S =
\begin{cases}
\emph{NAS}, & (\exists k:\,\phi_k > \gamma)\;\lor\;\bigl(\theta > \alpha \land y_I < y_{\mathrm{ank}}\bigr),\\[6pt]
\emph{AS},     & (\forall k:\,\phi_k \le \gamma)\;\land\;\bigl(\theta \le \alpha \lor y_I \ge y_{\mathrm{ank}}\bigr).
\end{cases}
\end{equation}

\begin{figure}[t]
\vspace{0.2cm}
    \centering
    \includegraphics[width=0.99\linewidth]{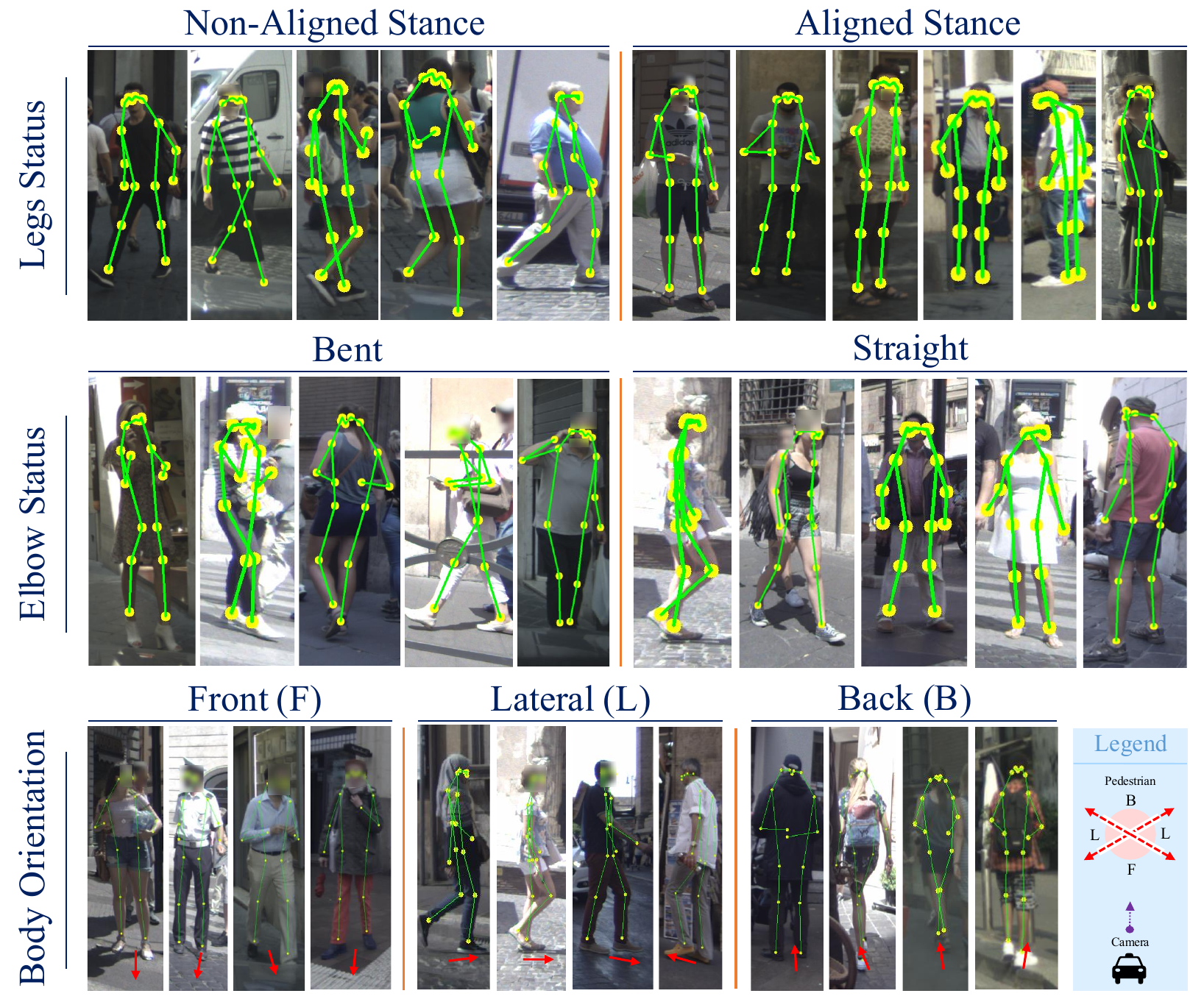}
    \caption{Illustration of random instances from pedestrian legs status, elbow status, and body orientation.}
    \label{fig:attribute_sample}
\end{figure}

 An aligned stance typically corresponds to a stationary pedestrian, whereas a non-aligned stance approximates walking. As only single images are available, this method can capture movement status for our analysis. Samples of automatically labeled pedestrian crops for each attribute, verified by random visual checks, are shown in Fig.~\ref{fig:attribute_sample}.

\subsubsection{Elbow Status}

Let \(S_k,E_k,W_k\) be the shoulder, elbow, and wrist points for side \(k\in\{\text{left},\text{right}\}\). Using the angle definition from \eqref{eq:angle_eq}, we compute: $ \psi_k = \measuredangle(E_k - S_k,\;W_k - E_k)$.

The elbow label \(L\) is determined by:
\begin{equation}
L = \begin{cases}
    \text{Bent elbow}, & \max\bigl\{\psi_{\text{left}},\, \psi_{\text{right}}\bigr\} \geq 90^\circ, \\
    \text{Straight elbow}, & \max\bigl\{\psi_{\text{left}},\, \psi_{\text{right}}\bigr\} < 90^\circ.
\end{cases}
\end{equation}
Bent elbows often correspond to holding a phone, carrying items, pushing a cart, gesturing, or manipulating objects.

\subsubsection{Occlusion Status}

We analyze occlusion impacts using per-joint visibility labels from ECP-DP annotations. Each joint forms a separate subgroup for analysis (17 total), where a joint is considered \emph{occluded} if hidden by objects, pedestrians, or self-occlusion and \emph{visible} otherwise. Detection rates are compared between these visibility states for each joint subgroup to identify which occlusions most degrade model performance.


\subsubsection{Body Orientation (view)}

Each pedestrian orientation is provided in the annotations as a scalar value \( a \in [0^\circ, 360^\circ) \), where \( a = 0^\circ \) indicates a pedestrian directly facing the ego-vehicle. We categorize orientations into three classes:

\begin{equation}
\text{View} =
\begin{cases}
\text{front}, & (0^\circ \leq a \leq 60^\circ) \lor (300^\circ \leq a < 360^\circ), \\[4pt]
\text{lateral}, & (60^\circ < a < 120^\circ) \lor (240^\circ < a < 300^\circ), \\[4pt]
\text{back}, & (120^\circ \leq a \leq 240^\circ).
\end{cases}
\end{equation}

Front- and back-facing pedestrians share similar silhouettes but differ in face visibility and clothing patterns. Lateral views are safety-critical as they typically represent pedestrians crossing the road perpendicular to the vehicle's path. These side profiles require precise detection due to their higher collision risk and distinct body kinematics during motion.

\subsubsection{Instance Size}

Pedestrian height is not treated as a separate  fairness subgroup but used to verify consistency of pose-induced biases. We split instances at 110\,px into small (\(<110\)\,px) and large (\(\ge110\)\,px) groups of roughly equal size. Large pedestrians are often closer to the vehicle, making detection failures more hazardous, whereas small instances have less precise pose annotations due to limited visibility of body details. By repeating all of our evaluations on the large-instance subset, we check whether pose-induced biases persist regardless of pedestrian size in the image.

\subsection{Evaluation Metrics}  

A pedestrian detector generates predicted boxes $\{b_i\}$ with scores $s_i \in [0,1]$. To mitigate the effect of random guesses, we apply a confidence score threshold (CST) $\tau \in \{0, 0.25, 0.5, 0.75\}$, retaining only $b_i$ with $s_i \ge \tau$.  

Following the methodology in \cite{braun2019eurocity}, to compute true positives (TP), false positives (FP), and false negatives (FN), we perform greedy matching using Intersection over Union (IoU). For any detection $b$ and ground truth $g$,  
\begin{equation}  
\mathrm{IoU}(b,g) = \frac{\mathrm{area}(b \cap g)}{\mathrm{area}(b \cup g)}.  
\end{equation}  
Detections are sorted by $s_i$ descending. For each $b_i$, if $\max_j \mathrm{IoU}(b_i, g_j) \ge 0.5$, we match $b_i$ to the $g_j$ with the highest IoU (as TP) and mark all other $b_k$ overlapping that $g_j$ as FP. If no $g_j$ meets the threshold, $b_i$ is FP. Unmatched $g_j$ after all detections are FN.  

Let $\mathrm{TP}_c^\tau$ and $\mathrm{FN}_c^\tau$ be the true positive and false negative counts for subgroup $c$ at threshold $\tau$. The subgroup miss-rate (MR) for $c$ is  
\begin{equation}  
MR_c^\tau = \frac{\mathrm{FN}_c^\tau}{\mathrm{TP}_c^\tau + \mathrm{FN}_c^\tau},  
\end{equation}  
and the overall miss-rate is defined similarly over all instances.  

\subsection{Fairness Metric: EOD and Cohen’s \(h\)}  

We measure fairness using the Equal Opportunity Difference (EOD) \cite{hardt2016equality} at each CST $\tau$. For a binary attribute with subgroups $c_{\text{maj}}$ (majority) and $c_{\text{min}}$ (minority), we compute  
\begin{equation}  
\mathrm{EOD}^\tau = MR_{\,c_{\text{min}}}^\tau - MR_{\,c_{\text{maj}}}^\tau.  
\end{equation}  
EOD uses only true positive and false negative rates, which are available for each subgroup. Metrics requiring subgroup labels for false positives (e.g., AOD \cite{zhang2021ignorance}) are inapplicable here, as incorrect detections lack annotations (no actual person).  

To quantify effect size independently of sample count, we compute Cohen’s $h$:  
\begin{equation}  
\Theta_c^\tau = \sin^{-1}\!\bigl(\sqrt{MR_c^\tau}\bigr), \quad  
h^\tau = 2\bigl(\Theta_{\,c_{\text{min}}}^\tau - \Theta_{\,c_{\text{maj}}}^\tau\bigr).  
\end{equation}  
For large datasets ($N > 10^5$), even trivial miss-rate differences can yield statistically significant $p$-values. Cohen’s $h$ addresses this by quantifying disparity magnitude independently of $N$. We report $|h^\tau|$ to focus on disparity magnitude.



\section{EXPERIMENTS}
\subsection{Dataset}
The EuroCity Persons Dense Pose (ECP-DP) \cite{braun21simple} dataset provides frontal and two synchronized side camera views captured across 31 European cities in 12 countries. To maximize sample size, we combine the training and validation sets (34,720 images total), as test set annotations are unavailable. From these, 194,909 pedestrian instances with bounding boxes taller than 60 pixels include pose annotations: 17 annotated 2D keypoints with occlusion status and body orientation angles. We used the proposed method to automatically derive four attributes from this data: leg status, elbow status, per-joint occlusion patterns, and body orientation categories (front/lateral/back).

\begin{table}[tb]
\vspace{0.2cm}
\centering
\scriptsize
\caption{Distribution (\%) of attributes in the ECP‐DP (train+val).}
\label{tab:ecp_distribution}
\renewcommand{\arraystretch}{1.2}
\setlength{\tabcolsep}{1.8pt}
\begin{tabular}{cc|cc|ccc|cc}
\toprule
  \multicolumn{2}{c}{\textbf{Leg status}} &
  \multicolumn{2}{c}{\textbf{Elbow status}} &
  \multicolumn{3}{c}{\textbf{Body Orientation}} &
  \multicolumn{2}{c}{\textbf{Height (px)}}\\
\cmidrule(lr){1-2}\cmidrule(lr){3-4}\cmidrule(lr){5-7}\cmidrule(lr){8-9}
  Aligned & Non-aligned
 & Straight & Bent
 & Front & Lateral & Back
 & 60–110 & $>$110 \\
  \BarSparkA{38.9}
 & \BarSparkB{61.1}
 & \BarSparkA{74.7}
 & \BarSparkB{25.3}
 & \BarSparkA{38.6}
 & \BarSparkC{31.3}
 & \BarSparkB{30.1}
 & \BarSparkA{49.2}
 & \BarSparkB{50.8} \\
\bottomrule
\end{tabular}
\end{table}

\begin{table*}[th]
\vspace{0.2cm}
\centering
\caption{EOD (\%) and Cohen's $h$-values for joint occlusion impact for detectors and CST values. MR indicates the overall miss-rate (\%). As occlusion consistently raises miss-rates, absolute EODs are used for simplicity. Bold $h$-values $\ge0.6$ show high impact.}

\label{tab:res_joint}
\renewcommand{\arraystretch}{0.8}  
\setlength{\tabcolsep}{1.5pt}        
\scalebox{0.615}{                   
\begin{tabular}{@{}c|c|c||aa|cc|aa|cc|aa|cc||aa|cc|aa|cc|aa|cc||aa|cc|aa|cc|aa@{}}

\toprule 
\textbf{Algorithm} & \textbf{CST} & \textbf{Miss-rate} 
& \multicolumn{2}{a|}{\textbf{ankle\_left}} 
& \multicolumn{2}{c}{\textbf{ankle\_right}}
& \multicolumn{2}{a|}{\textbf{knee\_left}} 
& \multicolumn{2}{c|}{\textbf{knee\_right}} 
& \multicolumn{2}{a|}{\textbf{hip\_left}} 
& \multicolumn{2}{c|}{\textbf{hip\_right}} 

& \multicolumn{2}{a|}{\textbf{wrist\_left}}
& \multicolumn{2}{c|}{\textbf{wrist\_right}} 

& \multicolumn{2}{a|}{\textbf{elbow\_left}} 
& \multicolumn{2}{c|}{\textbf{elbow\_right}} 

& \multicolumn{2}{a|}{\textbf{shoulder\_left}} 
& \multicolumn{2}{c|}{\textbf{shoulder\_right}} 

& \multicolumn{2}{a|}{\textbf{ear\_left}} 
& \multicolumn{2}{c|}{\textbf{ear\_right}} 
& \multicolumn{2}{a|}{\textbf{eye\_left}} 
& \multicolumn{2}{c|}{\textbf{eye\_right}} 
& \multicolumn{2}{a|}{\textbf{nose}} \\

\cmidrule(lr){4-5}  \cmidrule(lr){6-7} \cmidrule(lr){8-9} \cmidrule(lr){10-11} \cmidrule(lr){12-13}\cmidrule(lr){14-15}\cmidrule(lr){16-17}\cmidrule(lr){18-19}\cmidrule(lr){20-21}\cmidrule(lr){22-23}\cmidrule(lr){24-25}\cmidrule(lr){26-27}\cmidrule(lr){28-29}\cmidrule(lr){30-31}\cmidrule(lr){32-33}\cmidrule(lr){34-35}\cmidrule(lr){36-37}
 & $\tau$ & MR & \textbf{EOD} & \textbf{h}  & \textbf{EOD} & \textbf{h} & \textbf{EOD} & \textbf{h}  & \textbf{EOD} & \textbf{h} & \textbf{EOD} & \textbf{h}  & \textbf{EOD} & \textbf{h} & \textbf{EOD} & \textbf{h}  & \textbf{EOD} & \textbf{h} & \textbf{EOD} & \textbf{h} & \textbf{EOD} & \textbf{h} & \textbf{EOD} & \textbf{h} & \textbf{EOD} & \textbf{h} & \textbf{EOD} & \textbf{h} & \textbf{EOD} & \textbf{h} & \textbf{EOD} & \textbf{h} & \textbf{EOD} & \textbf{h} & \textbf{EOD} & \textbf{h}  \\
\midrule

ALFNet & 0.0 & 7.17 & 7.8 & 0.29 & 7.5 & 0.28 & 7.9 & 0.29 & 7.7 & 0.28 & 6.6 & 0.25 & 6.5 & 0.25 & 4.2 & 0.16 & 4.1 & 0.16 & 5.3 & 0.20 & 5.3 & 0.20 & 5.5 & 0.20 & 5.6 & 0.20 & 2.8 & 0.11 & 2.8 & 0.11 & 1.5 & 0.06 & 1.6 & 0.06 & 1.4 & 0.05 \\
ALFNet & 0.25 & 34.11 & 28.7 & \textbf{0.60} & 27.6 & 0.58 & 28.4 & 0.59 & 27.5 & 0.57 & 21.0 & 0.44 & 20.4 & 0.43 & 13.7 & 0.29 & 13.1 & 0.28 & 15.6 & 0.33 & 15.5 & 0.33 & 15.7 & 0.33 & 15.1 & 0.31 & 6.6 & 0.14 & 7.3 & 0.15 & 4.5 & 0.10 & 4.9 & 0.10 & 4.4 & 0.09 \\
ALFNet & 0.50 & 43.94 & 31.7 & \textbf{0.65} & 30.6 & \textbf{0.62} & 30.9 & \textbf{0.63} & 30.0 & \textbf{0.61} & 22.4 & 0.45 & 22.0 & 0.44 & 14.5 & 0.29 & 14.2 & 0.29 & 16.2 & 0.33 & 16.4 & 0.33 & 15.8 & 0.32 & 15.5 & 0.31 & 6.4 & 0.13 & 7.4 & 0.15 & 4.4 & 0.09 & 5.0 & 0.10 & 4.2 & 0.09 \\
ALFNet & 0.75 & 60.62 & 30.1 & \textbf{0.65} & 29.4 & \textbf{0.63} & 28.5 & \textbf{0.61} & 27.9 & 0.59 & 19.7 & 0.41 & 19.6 & 0.41 & 13.4 & 0.27 & 13.2 & 0.27 & 14.6 & 0.30 & 15.1 & 0.31 & 13.7 & 0.29 & 13.9 & 0.29 & 5.1 & 0.10 & 6.2 & 0.13 & 4.0 & 0.08 & 4.6 & 0.09 & 3.5 & 0.07 \\
\midrule

Cascade R-CNN & 0.0 & 1.20 & 1.5 & 0.14 & 1.6 & 0.14 & 1.5 & 0.13 & 1.6 & 0.14 & 1.1 & 0.10 & 1.1 & 0.10 & 0.7 & 0.06 & 0.7 & 0.07 & 0.8 & 0.07 & 0.9 & 0.08 & 0.8 & 0.07 & 0.9 & 0.08 & 0.3 & 0.03 & 0.4 & 0.03 & 0.3 & 0.03 & 0.3 & 0.03 & 0.4 & 0.04 \\
Cascade R-CNN & 0.25 & 5.63 & 6.4 & 0.27 & 6.6 & 0.28 & 6.6 & 0.27 & 6.8 & 0.28 & 5.7 & 0.24 & 5.6 & 0.24 & 3.5 & 0.15 & 3.6 & 0.16 & 4.3 & 0.18 & 4.6 & 0.19 & 5.2 & 0.21 & 5.2 & 0.21 & 2.4 & 0.10 & 2.9 & 0.12 & 2.1 & 0.09 & 2.2 & 0.10 & 2.5 & 0.11 \\
Cascade R-CNN & 0.50 & 8.14 & 8.8 & 0.31 & 9.0 & 0.32 & 9.3 & 0.33 & 9.6 & 0.34 & 8.3 & 0.30 & 8.2 & 0.30 & 5.2 & 0.19 & 5.2 & 0.19 & 6.5 & 0.23 & 6.6 & 0.24 & 7.7 & 0.26 & 7.6 & 0.26 & 3.6 & 0.13 & 4.1 & 0.15 & 2.8 & 0.10 & 3.0 & 0.11 & 3.6 & 0.13 \\
Cascade R-CNN & 0.75 & 10.80 & 11.9 & 0.37 & 12.1 & 0.38 & 12.4 & 0.39 & 12.6 & 0.39 & 10.5 & 0.33 & 10.4 & 0.33 & 6.6 & 0.22 & 6.6 & 0.22 & 8.0 & 0.25 & 8.2 & 0.26 & 9.0 & 0.27 & 9.2 & 0.28 & 4.2 & 0.13 & 4.8 & 0.15 & 3.5 & 0.11 & 3.6 & 0.12 & 4.4 & 0.14 \\
\midrule

CSP & 0.0 & 12.61 & 7.9 & 0.23 & 7.6 & 0.22 & 8.2 & 0.24 & 7.9 & 0.23 & 7.1 & 0.21 & 7.0 & 0.21 & 4.7 & 0.14 & 4.6 & 0.14 & 6.1 & 0.18 & 6.2 & 0.18 & 7.0 & 0.20 & 6.7 & 0.20 & 3.6 & 0.11 & 3.7 & 0.11 & 1.8 & 0.06 & 2.1 & 0.07 & 1.5 & 0.04 \\
CSP & 0.25 & 25.14 & 16.8 & 0.38 & 16.4 & 0.37 & 17.6 & 0.40 & 17.3 & 0.39 & 15.1 & 0.34 & 15.1 & 0.34 & 9.7 & 0.22 & 9.9 & 0.23 & 12.2 & 0.28 & 12.7 & 0.29 & 13.6 & 0.30 & 13.3 & 0.30 & 6.9 & 0.16 & 7.4 & 0.17 & 3.5 & 0.08 & 4.1 & 0.09 & 2.8 & 0.07 \\
CSP & 0.50 & 35.77 & 24.0 & 0.50 & 23.6 & 0.49 & 24.6 & 0.51 & 24.1 & 0.50 & 20.7 & 0.43 & 20.8 & 0.43 & 13.2 & 0.28 & 13.6 & 0.28 & 16.2 & 0.34 & 17.2 & 0.36 & 17.1 & 0.35 & 17.4 & 0.36 & 7.9 & 0.16 & 9.3 & 0.19 & 4.4 & 0.09 & 5.4 & 0.11 & 3.7 & 0.08 \\
CSP & 0.75 & 52.23 & 30.2 & \textbf{0.62} & 29.9 & \textbf{0.61} & 30.4 & \textbf{0.62} & 29.8 & \textbf{0.61} & 24.9 & 0.51 & 25.2 & 0.51 & 15.9 & 0.32 & 16.3 & 0.33 & 19.1 & 0.39 & 20.4 & 0.41 & 18.8 & 0.38 & 19.7 & 0.40 & 8.1 & 0.16 & 10.0 & 0.20 & 4.1 & 0.08 & 5.5 & 0.11 & 2.8 & 0.06 \\
\midrule

MGAN & 0.0 & 15.08 & 15.5 & 0.42 & 15.1 & 0.41 & 15.2 & 0.41 & 14.9 & 0.40 & 10.7 & 0.29 & 10.5 & 0.29 & 7.2 & 0.20 & 6.8 & 0.19 & 7.9 & 0.22 & 7.7 & 0.21 & 6.7 & 0.18 & 6.5 & 0.17 & 2.0 & 0.06 & 2.1 & 0.06 & 2.0 & 0.06 & 2.0 & 0.06 & 2.0 & 0.06 \\
MGAN & 0.25 & 41.87 & 31.8 & \textbf{0.65} & 31.0 & \textbf{0.63} & 31.6 & \textbf{0.65} & 30.9 & \textbf{0.63} & 23.2 & 0.47 & 22.5 & 0.46 & 15.3 & 0.31 & 14.5 & 0.29 & 16.7 & 0.34 & 16.5 & 0.33 & 15.5 & 0.31 & 15.0 & 0.30 & 5.6 & 0.11 & 6.3 & 0.13 & 4.9 & 0.10 & 5.1 & 0.10 & 4.7 & 0.10 \\
MGAN & 0.50 & 44.36 & 32.4 & \textbf{0.66} & 31.7 & \textbf{0.64} & 32.2 & \textbf{0.66} & 31.5 & \textbf{0.64} & 23.6 & 0.48 & 22.9 & 0.46 & 15.5 & 0.31 & 14.6 & 0.30 & 17.0 & 0.34 & 16.7 & 0.34 & 15.8 & 0.32 & 15.4 & 0.31 & 5.6 & 0.11 & 6.3 & 0.13 & 4.9 & 0.10 & 5.1 & 0.10 & 4.7 & 0.10 \\
MGAN & 0.75 & 46.98 & 32.7 & \textbf{0.67} & 32.0 & \textbf{0.65} & 32.5 & \textbf{0.66} & 31.8 & \textbf{0.65} & 23.7 & 0.48 & 23.1 & 0.47 & 15.7 & 0.32 & 14.8 & 0.30 & 17.1 & 0.34 & 16.9 & 0.34 & 15.9 & 0.32 & 15.5 & 0.31 & 5.7 & 0.11 & 6.3 & 0.13 & 4.9 & 0.10 & 5.1 & 0.10 & 4.8 & 0.10 \\
\midrule

F2DNet & 0.0 & 3.57 & 3.5 & 0.18 & 3.4 & 0.17 & 3.5 & 0.18 & 3.5 & 0.18 & 3.4 & 0.18 & 3.3 & 0.17 & 2.1 & 0.11 & 2.0 & 0.11 & 2.8 & 0.15 & 2.8 & 0.15 & 3.4 & 0.17 & 3.3 & 0.17 & 1.7 & 0.09 & 1.8 & 0.10 & 1.1 & 0.06 & 1.0 & 0.06 & 1.1 & 0.06 \\
F2DNet & 0.25 & 11.42 & 11.7 & 0.36 & 11.8 & 0.36 & 12.2 & 0.37 & 12.1 & 0.37 & 10.6 & 0.33 & 10.4 & 0.32 & 6.7 & 0.21 & 6.5 & 0.21 & 8.2 & 0.25 & 8.3 & 0.26 & 9.3 & 0.28 & 9.1 & 0.27 & 4.3 & 0.13 & 4.6 & 0.14 & 3.6 & 0.12 & 3.7 & 0.12 & 4.2 & 0.13 \\
F2DNet & 0.50 & 20.06 & 21.2 & 0.52 & 21.1 & 0.52 & 21.6 & 0.52 & 21.3 & 0.52 & 17.4 & 0.43 & 17.3 & 0.43 & 11.3 & 0.28 & 11.2 & 0.28 & 13.1 & 0.32 & 13.4 & 0.33 & 13.6 & 0.33 & 13.7 & 0.33 & 5.8 & 0.14 & 6.8 & 0.17 & 5.6 & 0.14 & 6.0 & 0.15 & 6.5 & 0.16 \\
F2DNet & 0.75 & 39.50 & 36.7 & \textbf{0.76} & 36.6 & \textbf{0.76} & 35.6 & \textbf{0.73} & 35.4 & \textbf{0.73} & 25.9 & 0.53 & 26.5 & 0.54 & 17.1 & 0.35 & 17.4 & 0.36 & 18.4 & 0.38 & 19.4 & 0.40 & 17.1 & 0.35 & 17.9 & 0.36 & 6.0 & 0.12 & 8.3 & 0.17 & 6.5 & 0.13 & 7.7 & 0.16 & 7.5 & 0.15 \\
\midrule

YOLOv12-S (gen) & 0.0 & 5.92 & 13.2 & 0.56 & 12.8 & 0.54 & 13.4 & 0.56 & 13.1 & 0.55 & 10.1 & 0.42 & 9.9 & 0.42 & 6.1 & 0.27 & 6.1 & 0.27 & 6.0 & 0.25 & 6.2 & 0.26 & 4.5 & 0.18 & 4.6 & 0.18 & 1.8 & 0.07 & 2.1 & 0.09 & 1.3 & 0.06 & 1.6 & 0.07 & 1.8 & 0.08 \\
YOLOv12-S (gen) & 0.25 & 19.79 & 32.7 & \textbf{0.81} & 32.4 & \textbf{0.80} & 33.6 & \textbf{0.83} & 33.3 & \textbf{0.82} & 25.8 & \textbf{0.64} & 25.7 & \textbf{0.64} & 16.0 & 0.41 & 16.0 & 0.41 & 16.8 & 0.41 & 17.1 & 0.42 & 14.7 & 0.35 & 14.9 & 0.36 & 5.5 & 0.14 & 6.6 & 0.16 & 4.1 & 0.10 & 4.9 & 0.12 & 5.6 & 0.14 \\
YOLOv12-S (gen) & 0.50 & 31.26 & 39.1 & \textbf{0.84} & 38.8 & \textbf{0.84} & 39.8 & \textbf{0.85} & 39.6 & \textbf{0.85} & 30.1 & \textbf{0.65} & 29.9 & \textbf{0.65} & 19.2 & 0.42 & 19.1 & 0.42 & 20.7 & 0.44 & 21.1 & 0.45 & 18.5 & 0.39 & 18.7 & 0.39 & 6.6 & 0.14 & 7.6 & 0.16 & 4.9 & 0.11 & 5.6 & 0.12 & 6.4 & 0.14 \\
YOLOv12-S (gen) & 0.75 & 61.96 & 33.1 & \textbf{0.73} & 32.7 & \textbf{0.71} & 32.8 & \textbf{0.72} & 32.3 & \textbf{0.70} & 23.5 & 0.50 & 23.3 & 0.49 & 17.4 & 0.36 & 16.3 & 0.34 & 18.8 & 0.40 & 18.6 & 0.39 & 16.9 & 0.36 & 16.6 & 0.35 & 5.4 & 0.11 & 5.5 & 0.12 & 4.7 & 0.10 & 5.2 & 0.11 & 4.9 & 0.10 \\
\midrule

YOLOv12-X (gen) & 0.0 & 5.52 & 12.4 & 0.54 & 12.1 & 0.53 & 12.5 & 0.54 & 12.2 & 0.53 & 9.3 & 0.40 & 9.2 & 0.41 & 5.8 & 0.26 & 5.7 & 0.26 & 5.7 & 0.24 & 5.7 & 0.25 & 4.3 & 0.18 & 4.4 & 0.18 & 1.6 & 0.07 & 2.0 & 0.08 & 1.2 & 0.05 & 1.4 & 0.06 & 1.7 & 0.07 \\
YOLOv12-X (gen) & 0.25 & 18.90 & 32.9 & \textbf{0.83} & 32.6 & \textbf{0.82} & 33.7 & \textbf{0.84} & 33.5 & \textbf{0.84} & 25.5 & \textbf{0.64} & 25.3 & \textbf{0.64} & 16.0 & 0.42 & 15.9 & 0.42 & 16.2 & 0.41 & 16.5 & 0.42 & 13.7 & 0.33 & 13.9 & 0.34 & 5.1 & 0.13 & 5.9 & 0.15 & 4.1 & 0.10 & 4.4 & 0.11 & 5.6 & 0.14 \\
YOLOv12-X (gen) & 0.50 & 27.49 & 39.0 & \textbf{0.87} & 38.7 & \textbf{0.86} & 39.5 & \textbf{0.88} & 39.5 & \textbf{0.88} & 29.4 & \textbf{0.66} & 29.2 & \textbf{0.65} & 18.6 & 0.42 & 18.5 & 0.42 & 19.4 & 0.43 & 19.8 & 0.44 & 16.9 & 0.37 & 17.1 & 0.37 & 5.7 & 0.13 & 6.8 & 0.15 & 4.9 & 0.11 & 5.3 & 0.12 & 6.6 & 0.15 \\
YOLOv12-X (gen) & 0.75 & 51.31 & 38.7 & \textbf{0.81} & 38.5 & \textbf{0.80} & 38.2 & \textbf{0.79} & 38.1 & \textbf{0.79} & 26.7 & 0.54 & 26.6 & 0.54 & 19.2 & 0.39 & 18.8 & 0.38 & 20.2 & 0.41 & 20.5 & 0.41 & 17.8 & 0.36 & 17.8 & 0.36 & 5.2 & 0.10 & 6.5 & 0.13 & 5.2 & 0.10 & 5.9 & 0.12 & 6.2 & 0.12 \\
\midrule

YOLOv12-S (fine) & 0.0 & 3.44 & 2.9 & 0.15 & 2.9 & 0.15 & 2.6 & 0.14 & 2.8 & 0.15 & 2.2 & 0.12 & 2.4 & 0.13 & 1.3 & 0.07 & 1.4 & 0.07 & 1.6 & 0.09 & 1.9 & 0.10 & 1.8 & 0.09 & 1.9 & 0.10 & 1.0 & 0.05 & 1.0 & 0.06 & 0.3 & 0.02 & 0.5 & 0.03 & 0.2 & 0.01 \\
YOLOv12-S (fine) & 0.25 & 22.94 & 20.9 & 0.48 & 20.9 & 0.49 & 20.4 & 0.47 & 20.5 & 0.48 & 16.4 & 0.39 & 16.6 & 0.39 & 10.0 & 0.24 & 10.3 & 0.25 & 12.3 & 0.29 & 13.0 & 0.31 & 12.7 & 0.29 & 13.1 & 0.30 & 6.0 & 0.14 & 7.0 & 0.17 & 3.5 & 0.08 & 4.3 & 0.10 & 3.7 & 0.09 \\
YOLOv12-S (fine) & 0.50 & 33.58 & 27.7 & 0.58 & 27.8 & 0.58 & 26.9 & 0.56 & 26.9 & 0.56 & 21.5 & 0.45 & 21.7 & 0.46 & 13.0 & 0.28 & 13.4 & 0.28 & 15.7 & 0.33 & 16.7 & 0.35 & 15.5 & 0.32 & 16.4 & 0.34 & 6.9 & 0.15 & 8.6 & 0.18 & 3.9 & 0.08 & 4.9 & 0.10 & 4.0 & 0.09 \\
YOLOv12-S (fine) & 0.75 & 56.82 & 31.1 & \textbf{0.65} & 31.3 & \textbf{0.66} & 30.1 & \textbf{0.63} & 29.8 & \textbf{0.62} & 23.8 & 0.49 & 24.2 & 0.50 & 14.3 & 0.29 & 14.8 & 0.30 & 17.1 & 0.35 & 18.3 & 0.37 & 16.6 & 0.34 & 17.6 & 0.36 & 6.9 & 0.14 & 8.9 & 0.18 & 3.1 & 0.06 & 4.4 & 0.09 & 2.4 & 0.05 \\
\midrule

Average & 0.25 & 22.47 & 22.7 & 0.55 & 22.4 & 0.54 & 23.0 & 0.55 & 22.7 & 0.55 & 17.9 & 0.44 & 17.7 & 0.43 & 11.4 & 0.28 & 11.2 & 0.28 & 12.8 & 0.31 & 13.0 & 0.32 & 12.6 & 0.30 & 12.5 & 0.30 & 5.3 & 0.13 & 6.0 & 0.15 & 3.8 & 0.10 & 4.2 & 0.11 & 4.2 & 0.11  \\

\bottomrule

\end{tabular}}
\end{table*}

\subsection{Subgroups Distributions}

\begin{figure}[tb]
    \vspace{-0.12cm}
    \centering
    \includegraphics[width=0.9\linewidth]{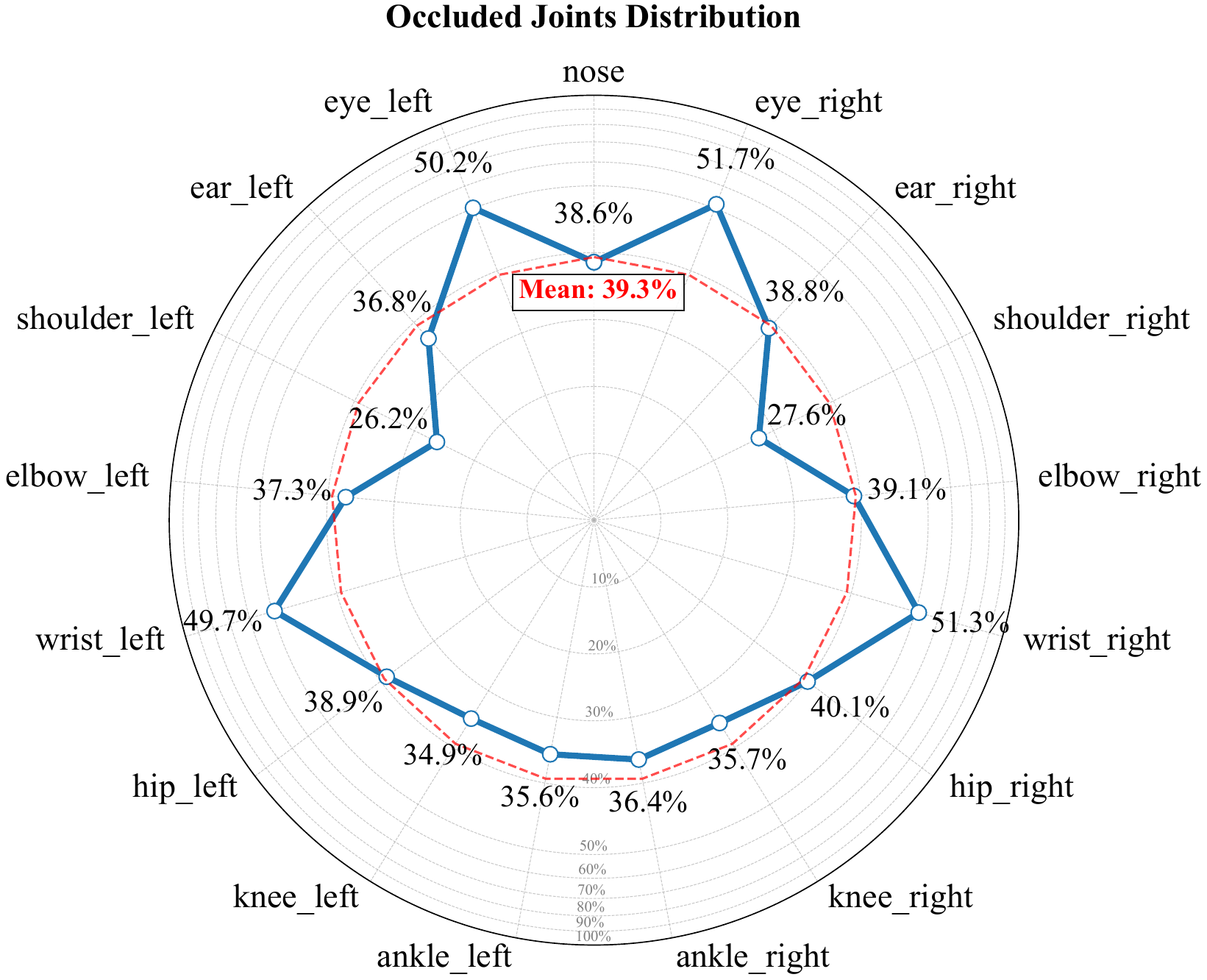}
    \caption{Per-joint occlusion rates (percentage) across body keypoints in the merged ECP-DP training/validation set. The red dashed line indicates the average occlusion rate (39.3\%) across all joints.}
    \label{fig:joint-occlusion-distr}
\end{figure}

We analyze the distribution of key attributes in the ECP-DP dataset (training and validation sets merged). Fig.~\ref{fig:joint-occlusion-distr} shows the per-joint occlusion distribution. Wrists and eyes are among the most frequently occluded joints, appearing occluded in nearly 50\% of the annotated pedestrians, while shoulders are the least occluded. Table~\ref{tab:ecp_distribution} summarizes the distributions of the three selected attributes and bounding box height. Most pedestrians exhibit moving leg stances (non-aligned) and straight elbows. Most pedestrians face frontally, with lateral views slightly outnumbering back-facing ones, although both remain less common than frontal orientations.

\subsection{Detection Methods}
In total, we evaluate eight detectors, chosen for their diverse architectures and varying degrees of specialization in pedestrian detection. YOLOv12-S~\cite{yolov12} and YOLOv12-X~\cite{yolov12} are general-purpose object detectors, pre-trained on MS-COCO~\cite{coco} and without any pedestrian-specific fine-tuning. The S variant uses a lightweight backbone optimized for speed, while the X variant employs a deeper architecture with higher representational capacity. We additionally fine-tune YOLOv12-S on the CityPersons~\cite{cityperson} dataset for 25 epochs to better adapt them to pedestrian detection. These are referred to as YOLOv12 (gen) for the general models and YOLOv12 (fine) for the fine-tuned one. Also, we choose five pedestrian-specific models—ALFNet~\cite{alfnet}, Cascade R-CNN (with HRNet as the main backbone) from Pedestron~\cite{pedestron}, CSP~\cite{csp}, MGAN~\cite{mgan}, and F2DNet~\cite{f2dnet}—all pre-trained on CityPersons~\cite{cityperson} and tailored for pedestrian detection.

\subsection{Results}

\subsubsection{Impact of Joint Occlusions}

Table~\ref{tab:res_joint} (with absolute EOD values for readability) shows that occluding any joint significantly raises miss‐rate (\(p<0.01\)). Joints are grouped into lower body (ankle, knee, hip), upper body (wrist, elbow, shoulder), and head (eye, ear, nose). At \(\tau=0.25\), mean EOD is 21.1\% for lower body, 12.2\% for upper body, and 4.7\% for head. Detectors struggle most when lower‐body joints are occluded. A typical case is pedestrians behind low, wide obstacles (e.g., vehicles), leaving only upper body and head visible, which reduces detection performance.

\subsubsection{Impact of Leg Status}

Across all detectors and CSTs, pedestrians with aligned stances have significantly higher miss‐rates (\(p<0.01\); see Table~\ref{tab:res_attr}). Non-aligned samples (open legs or bent knees) are more prevalent in ECP-DP, yielding positive EOD values. Since aligned stances often indicate stationary pedestrians at crossings or lights, their lower detection rate might pose safety risks.

\begin{table}[tb]
\vspace{0.0cm}
\centering
\caption{\scriptsize EOD (\%) and Cohen’s h values for pose-related attributes across detectors. Bold values ($h\ge0.2$) indicate medium to high impact.}
\label{tab:res_attr}
\renewcommand{\arraystretch}{0.75}  
\setlength{\tabcolsep}{1.5pt}        
\scalebox{0.67}{                   
\begin{tabular}{@{}c|c|c|cc|cc|cccccc@{}}

\toprule 
\textbf{Algorithm} & \textbf{CST} & \textbf{Miss-rate} 
& \multicolumn{2}{c|}{\textbf{Legs}} 
& \multicolumn{2}{c|}{\textbf{Elbow}} 
& \multicolumn{6}{c}{\textbf{Body Orientation}}  \\

\cmidrule(lr){4-5}  \cmidrule(lr){6-7} \cmidrule(lr){8-13} 
 & $\tau$ & MR & $\mathrm{EOD}$ & $\mathrm{h}$  & $\mathrm{EOD}$ & $\mathrm{h}$ & $\mathrm{EOD}_{F-L}$ & $\mathrm{h}_{F-L}$  & $\mathrm{EOD}_{L-B}$ & $\mathrm{h}_{L-B}$ & $\mathrm{EOD}_{F-B}$ & $\mathrm{h}_{F-B}$  \\
\midrule

ALFNet & 0.0  & 7.17 & $+2.0$ & $0.07$  & $-1.9$ & $0.08$   & $+2.4$ & $0.09$ &  $-3.7$ & $0.14$ & $-1.3$ & $0.05$ \\
ALFNet & 0.25 & 34.11 & $+8.0$ & $0.17$  & $-5.7$ & $0.12$   & $+5.5$ & $0.11$ &  $-9.8$ & \textbf{0.21} & $-4.3$ & $0.09$ \\
ALFNet & 0.50 & 43.94 & $+9.0$ & $0.18$  & $-5.9$ & $0.12$   & $+5.8$ & $0.12$ &  $-10.1$ & \textbf{0.20} & $-4.3$ & $0.09$ \\
ALFNet & 0.75 & 60.62 & $+9.2$ & $0.19$  & $-5.5$ & $0.11$   & $+5.9$ & $0.12$ &  $-9.2$ & $0.19$ & $-3.3$ & $0.07$ \\
\midrule

Cascade R-CNN & 0.0 & 1.20 &  $+0.6$ & $0.05$  & $-0.5$ & $0.05$  &  $-0.2$ & $0.02$ &  $-0.4$ & $0.04$ & $-0.6$ & $0.06$\\
Cascade R-CNN & 0.25 & 5.63 & $+1.7$ & $0.07$  & $-2.1$ & $0.10$  &  $+0.8$ & $0.03$ &  $-2.5$ & $0.11$ & $-1.7$ & $0.08$\\
Cascade R-CNN & 0.50 & 8.14 & $+2.1$ & $0.08$  & $-2.8$ & $0.11$  &  $+1.4$ & $0.05$ &  $-3.6$ & $0.13$ & $-2.2$ & $0.09$\\
Cascade R-CNN & 0.75 & 10.8 & $+2.9$ & $0.09$  & $-3.4$ & $0.11$  &  $+1.4$ & $0.03$ &  $-4.1$ & $0.14$ & $-2.7$ & $0.09$\\
\midrule

CSP & 0.0  & 12.61 &  $+0.9$ & $0.03$  & $-1.9$ & $0.06$  &  $+3.6$ & $0.10$ &  $-5.8$ & $0.18$ & $-2.2$ & $0.07$ \\
CSP & 0.25 & 25.14 &  $+2.1$ & $0.05$  & $-3.3$ & $0.08$  &  $+6.7$ & $0.15$ &  $-10.1$ & \textbf{0.23} & $-3.4$ & $0.08$ \\
CSP & 0.50 & 35.77 &  $+4.0$ & $0.08$  & $-4.4$ & $0.09$  &  $+8.1$ & $0.17$ &  $-12.0$ & \textbf{0.25} & $-3.9$ & $0.08$\\
CSP & 0.75 & 52.23 &  $+5.4$ & $0.11$  & $-5.0$ & $0.10$  &  $+10.0$ & \textbf{0.20} &  $-13.6$ & \textbf{0.27} & $-3.6$ & $0.07$ \\
\midrule

MGAN & 0.0  & 15.08 &  $+6.5$  & $0.18$  & $-4.1$ & $0.12$  &  $+1.4$ & $0.04$ &  $-4.4$ & $0.12$ & $-3.0$ & $0.09$     \\
MGAN & 0.25 & 41.87 &  $+10.5$ & \textbf{0.21}  & $-7.4$ & $0.15$  &  $+5.0$ & $0.10$ &  $-9.2$ & $0.19$ & $-4.2$ & $0.09$    \\
MGAN & 0.50 & 44.36 &  $+10.7$ & \textbf{0.21}  & $-7.5$ & $0.15$  &  $+5.2$ & $0.10$ &  $-9.3$ & $0.19$ & $-4.1$ & $0.08$    \\
MGAN & 0.75 & 46.98 &  $+10.8$ & \textbf{0.22}  & $-7.4$ & $0.15$  &  $+5.4$ & $0.11$ &  $-9.4$ & $0.19$ & $-4.0$ & $0.08$    \\

\midrule
F2DNet & 0.0  & 3.57 &  $+0.9$  & $0.05$  & $-1.3$ & $0.07$  & $+1.2$ & $0.06$ &  $-2.2$ & $0.12$ & $-1.0$ & $0.06$      \\
F2DNet & 0.25 & 11.42 &  $+2.9$ & $0.09$  & $-3.2$ & $0.10$  & $+2.4$ & $0.07$ &  $-4.8$ & $0.15$ & $-2.4$ & $0.08$     \\
F2DNet & 0.50 & 20.06 &  $+5.6$ & $0.14$  & $-5.0$ & $0.13$  & $+3.6$ & $0.09$ &  $-6.2$ & $0.16$ & $-2.6$ & $0.07$     \\
F2DNet & 0.75 & 39.50 &  $+10.8$ & \textbf{0.22}  & $-7.3$ & $0.15$  & $+5.8$ & $0.12$ &  $-7.7$ & $0.16$ & $-1.9$ & $0.04$     \\
\midrule

YOLOv12-S (gen) & 0.0  & 5.92 &  $+4.8$  & \textbf{0.20}  & $-2.4$ & $0.11$  & $+0.1$ & $0.00$ &  $-1.1$ & $0.05$ & $-1.0$ & $0.04$      \\
YOLOv12-S (gen) & 0.25 & 19.79 &  $+11.4$ & \textbf{0.28}  & $-7.1$ & $0.19$  & $+0.3$ & $0.01$ &  $-3.7$ & $0.09$ & $-3.3$ & $0.09$     \\
YOLOv12-S (gen) & 0.50 & 31.26 &  $+13.8$ & \textbf{0.30}  & $-9.1$ & \textbf{0.20}  & $+1.4$ & $0.03$ &  $-6.0$ & $0.13$ & $-4.6$ & $0.10$     \\
YOLOv12-S (gen) & 0.75 & 61.96 &  $+12.6$ & \textbf{0.26}  & $-7.7$ & $0.16$  & $+4.7$ & $0.10$ &  $-7.7$ & $0.16$ & $-3.0$ & $0.06$     \\
\midrule

YOLOv12-X (gen) & 0.0  & 5.52 &  $+4.6$  & \textbf{0.20}  & $-2.2$ & $0.10$  & $-0.3$ & $0.01$ &  $-0.7$ & $0.03$ & $-1.0$ & $0.05$      \\
YOLOv12-X (gen) & 0.25 & 18.90 &  $+12.1$ & \textbf{0.30}  & $-7.0$ & $0.19$  & $-0.7$ & $0.02$ &  $-2.8$ & $0.07$ & $-3.5$ & $0.09$     \\
YOLOv12-X (gen) & 0.50 & 27.49 &  $+14.8$ & \textbf{0.33}  & $-8.8$ & \textbf{0.20}  & $ 0.0$ & $0.00$ &  $-4.3$ & $0.10$ & $-4.3$ & $0.10$     \\
YOLOv12-X (gen) & 0.75 & 51.31 &  $+15.7$ & \textbf{0.32}  & $-9.1$ & $0.18$  & $+3.4$ & $0.07$ &  $-6.6$ & $0.13$ & $-3.3$ & $0.07$     \\
\midrule

YOLOv12-S (fine) & 0.0  & 3.44 &  $+0.3$  & $0.02$  & $-0.5$ & $0.03$  & $+0.7$ & $0.04$ &  $-1.6$ & $0.09$ & $-0.9$ & $0.05$      \\
YOLOv12-S (fine) & 0.25 & 22.94 &  $+5.4$ & $0.13$  & $-3.2$ & $0.08$  & $+4.0$ & $0.09$ &  $-7.9$ & $0.19$ & $-3.9$ & $0.10$     \\
YOLOv12-S (fine) & 0.50 & 33.58 &  $+6.8$ & $0.14$  & $-3.7$ & $0.08$  & $+5.2$ & $0.11$ &  $-9.8$ & \textbf{0.21} & $-4.6$ & $0.10$     \\
YOLOv12-S (fine) & 0.75 & 56.82 &  $+6.8$ & $0.14$  & $-4.3$ & $0.09$  & $+7.2$ & $0.15$ &  $-11.5$ & \textbf{0.23} & $-4.3$ & $0.09$     \\
\midrule

Average & 0.25 & 22.47 & $+6.8$ & 0.16 & $-4.9$ & 0.12 & $+3.0$ & 0.07 & $-6.3$ & 0.15 & $-3.3$ & 0.09 \\ 

\bottomrule
\end{tabular}}
\end{table}

\subsubsection{Impact of Elbow Status}

Bent elbows are common when pedestrians hold phones, carry items or manipulate objects, and these cases are detected more reliably. Although straight‐elbow samples are three times more frequent, their miss‐rate is higher (negative EOD). This gap is significant across all detectors (\(p<0.01\)) and suggests weaker detection for neutral arm postures, which may raise societal concern.

\subsubsection{Impact of Body Orientation}

Both the front–lateral EOD (positive) and the lateral–back EOD (negative) are statistically significant (\(p<0.01\)). Table~\ref{tab:res_attr} shows that laterally viewed pedestrians have higher miss-rates than front- or back-facing ones. Back-facing pedestrians are detected more reliably than front-facing ones. Laterally viewed pedestrians are often orthogonal to the vehicle and crossing the road, making them particularly safety-critical cases for AVs.

\subsubsection{Impact of Instance Size}

All experiments were repeated on instances with height \(>110\)\,px, excluding small samples. Here “full” denotes the entire dataset, and “large” only medium-to-large instances. The sign of every EOD bias remains unchanged. At \(\tau=0.25\), the mean difference across detectors, \(\Delta = \overline{\mathrm{EOD}}_{\text{full}} - \overline{\mathrm{EOD}}_{\text{large}}\), is $+0.11$ for leg status and $–0.62$ for elbow status. For orientation, \(\Delta_{F\!-\!L}=-1.21\), \(\Delta_{L\!-\!B}=-1.42\), and \(\Delta_{F\!-\!B}=-0.22\). A positive \(\Delta\) means tiny samples increase measured bias; a negative \(\Delta\) means bias is higher in larger instances. These small \(\Delta\) values and consistent signs for EODs in both cases confirm that instance size has a negligible effect on pose-induced fairness. The same holds for joint-occlusion analyses across body regions.

\subsubsection{Detectors Comparison}

\begin{figure}[tb]
    \vspace{0.2cm}
    \centering
    \includegraphics[width=1.0\linewidth]{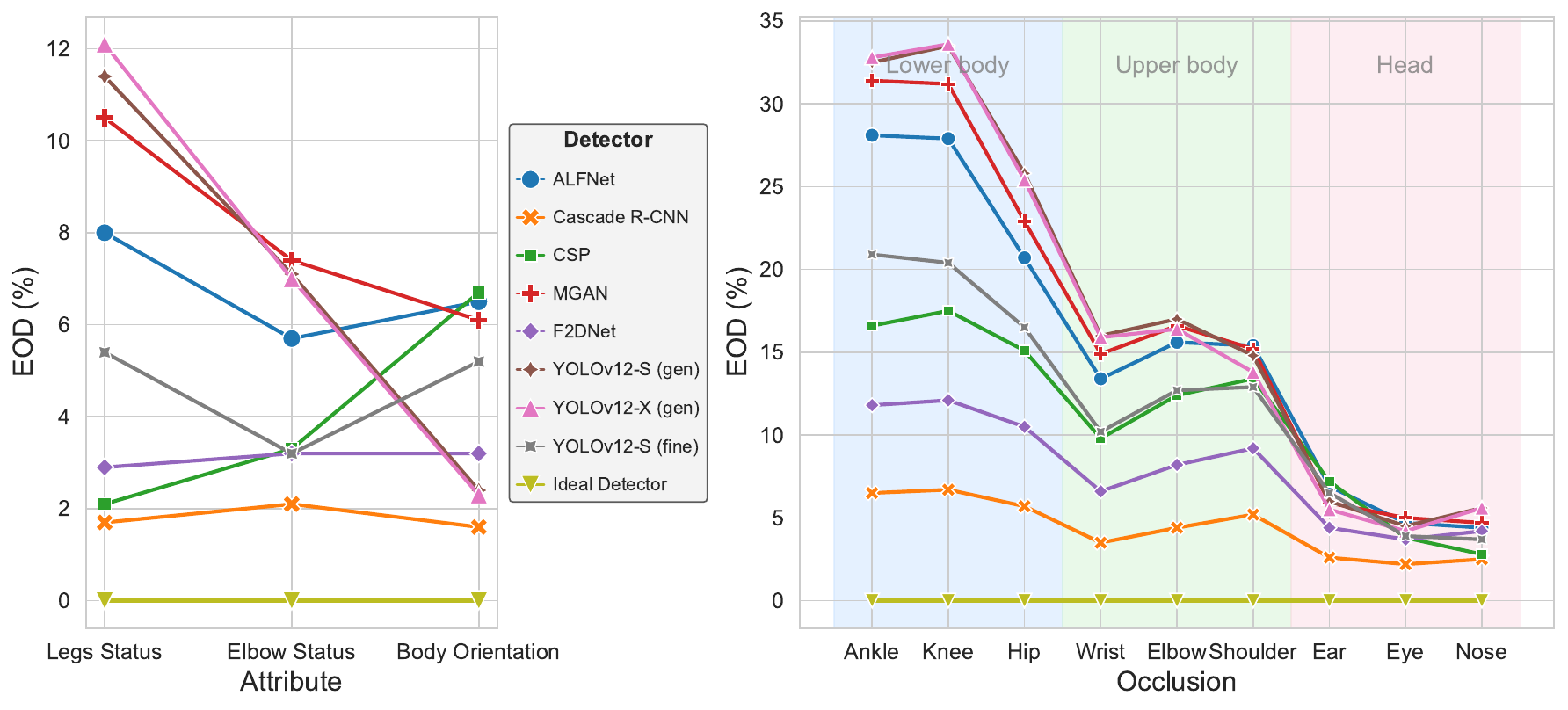}
    \caption{Visual comparison of all detectors across attributes and joint occlusion ($\tau=0.25$). For body orientation, three EOD values are averaged. EOD values for left and right joints are also averaged due to similar values.}
    \label{fig:detector_comparison}
\end{figure}

Cascade R-CNN achieves the lowest overall miss‐rate and the smallest average EOD across all attributes, with every \(\mathrm{EOD}<1.0\%\) at \(\tau=0.0\) (except on lower body occlusions). F2DNet is the next best performer. Both models sustain high recall as \(\tau\) increases, indicating greater confidence in their predictions, whereas other detectors suffer steep performance drops. In general, detectors with lower miss‐rates and higher confidence scores exhibit smaller biases. Among the YOLOv12 general-purpose (gen) variants, the X model shows a marginally lower miss‐rate than S. After fine-tuning, YOLOv12-S (fine) not only reduces miss‐rate (at $\tau=0.0$) but also yields significantly improved fairness for leg status, elbow status, and joint occlusions. Fig. \ref{fig:detector_comparison} shows a complete comparison of all detectors at $\tau=0.25$.

\section{Discussion}
Our analysis leverages a significantly larger sample set than earlier studies \cite{li2024bias}, which leads to more robust and generalizable findings. A minor limitation is that our labeling strategy based on 2D pose data might misclassify a few bent joints when the bend is along the depth axis. This could be improved by using 3D poses in future work. Although the EOD metric effectively quantifies subgroup disparities, its binary nature requires confidence score discretization and separate tests. 
Besides, even small detection biases are important to address because AVs are used at large scale, especially in cases involving occlusion or lateral views.

\section{CONCLUSION}
In this study, we present the fairness evaluations of eight different detectors, which reveal several pose-induced biases in pedestrian detection in the AD domain. Pedestrians with aligned legs ($\mathrm{EOD} = +6.8\%$) and straight elbows ($\mathrm{EOD} = -4.9\%$) face elevated miss-rates. Lateral views (critical for crossing scenarios) show the highest risk ($\mathrm{EOD} = -6.3\%$ for lateral-back and $\mathrm{EOD} = +3.0\%$ for front-lateral). 
Lower body occlusions degrade detection $1.73\times$ more than upper-body joints ($\mathrm{EOD} = 21.1\%$ vs. $\mathrm{EOD} = 12.2\%$), with head occlusions being the least impactful. In addition, bias direction for all subgroups remained stable regardless of bounding-box height. Notably, fine-tuning general-purpose detectors on pedestrian data reduced bias for most attributes. These biases demand focused attention in detector development to ensure equitable performance across diverse pedestrian attributes.



\bibliographystyle{IEEEtran} 
\bibliography{ref}

\begin{thebibliography}{10}
\providecommand{\url}[1]{#1}
\csname url@samestyle\endcsname
\providecommand{\newblock}{\relax}
\providecommand{\bibinfo}[2]{#2}
\providecommand{\BIBentrySTDinterwordspacing}{\spaceskip=0pt\relax}
\providecommand{\BIBentryALTinterwordstretchfactor}{4}
\providecommand{\BIBentryALTinterwordspacing}{\spaceskip=\fontdimen2\font plus
\BIBentryALTinterwordstretchfactor\fontdimen3\font minus \fontdimen4\font\relax}
\providecommand{\BIBforeignlanguage}[2]{{%
\expandafter\ifx\csname l@#1\endcsname\relax
\typeout{** WARNING: IEEEtran.bst: No hyphenation pattern has been}%
\typeout{** loaded for the language `#1'. Using the pattern for}%
\typeout{** the default language instead.}%
\else
\language=\csname l@#1\endcsname
\fi
#2}}
\providecommand{\BIBdecl}{\relax}
\BIBdecl

\bibitem{adchallenges}
B.~Padmaja, C.~V. Moorthy, N.~Venkateswarulu, and M.~M. Bala, ``Exploration of issues, challenges and latest developments in autonomous cars,'' \emph{Journal of Big Data}, vol.~10, no.~1, p.~61, 2023.

\bibitem{chen2024fairness}
Z.~Chen, J.~M. Zhang, M.~Hort, M.~Harman, and F.~Sarro, ``Fairness testing: A comprehensive survey and analysis of trends,'' \emph{ACM Trans. Softw. Eng. Methodol.}, vol.~33, no.~5, pp. 1--59, 2024.

\bibitem{wilson2019predictive}
B.~Wilson, J.~Hoffman, and J.~Morgenstern, ``Predictive inequity in object detection,'' \emph{arXiv preprint arXiv:1902.11097}, 2019.

\bibitem{braun21simple}
M.~Braun, F.~B. Flohr, S.~Krebs, U.~Kreßel, and D.~M. Gavrila, ``Simple pair pose - pairwise human pose estimation in dense urban traffic scenes,'' in \emph{\iv}, 2021, pp. 1545--1552.

\bibitem{yolov12}
Y.~Tian, Q.~Ye, and D.~Doermann, ``Yolov12: Attention-centric real-time object detectors,'' \emph{arXiv preprint arXiv:2502.12524}, 2025.

\bibitem{modern}
J.~Cao, Y.~Pang, J.~Xie, F.~S. Khan, and L.~Shao, ``From handcrafted to deep features for pedestrian detection: A survey,'' \emph{IEEE Trans. Pattern Anal. Mach. Intell.}, vol.~44, no.~9, pp. 4913--4934, 2021.

\bibitem{f2dnet}
A.~H. Khan, M.~Munir, L.~van Elst, and A.~Dengel, ``{ F2DNet: Fast Focal Detection Network for Pedestrian Detection },'' in \emph{\icpr}, 2022, pp. 4658--4664.

\bibitem{mgan}
Y.~Pang, J.~Xie, M.~H. Khan, R.~M. Anwer, F.~S. Khan, and L.~Shao, ``Mask-guided attention network for occluded pedestrian detection,'' in \emph{\iccv}, 2019, pp. 4966--4974.

\bibitem{alfnet}
W.~Liu, S.~Liao, and W.~Hu, ``Efficient single-stage pedestrian detector by asymptotic localization fitting and multi-scale context encoding,'' \emph{IEEE Trans. Image Process.}, vol.~29, pp. 1413--1425, 2020.

\bibitem{csp}
W.~Liu, S.~Liao, W.~Ren, W.~Hu, and Y.~Yu, ``High-level semantic feature detection: A new perspective for pedestrian detection,'' in \emph{\cvpr}, 2019, pp. 5182--5191.

\bibitem{pedestron}
I.~Hasan, S.~Liao, J.~Li, S.~U. Akram, and L.~Shao, ``Generalizable pedestrian detection: The elephant in the room,'' in \emph{\cvpr}, 2021, pp. 11\,323--11\,332.

\bibitem{cityperson}
S.~Zhang, R.~Benenson, and B.~Schiele, ``Citypersons: A diverse dataset for pedestrian detection,'' in \emph{\cvpr}, 2017, pp. 4457--4465.

\bibitem{coco}
T.-Y. Lin, M.~Maire, S.~Belongie, J.~Hays, P.~Perona, D.~Ramanan, P.~Doll{\'a}r, and C.~L. Zitnick, ``Microsoft coco: Common objects in context,'' in \emph{\eccv}, 2014, pp. 740--755.

\bibitem{yolo}
J.~Redmon, S.~Divvala, R.~Girshick, and A.~Farhadi, ``You only look once: Unified, real-time object detection,'' in \emph{\cvpr}, 2016, pp. 779--788.

\bibitem{mehrabi2021survey}
N.~Mehrabi, F.~Morstatter, N.~Saxena, K.~Lerman, and A.~Galstyan, ``A survey on bias and fairness in machine learning,'' \emph{ACM CSUR}, vol.~54, no.~6, pp. 1--35, 2021.

\bibitem{alexey-IAVVC25}
A.~L. Flores~Comeca, N.~Masarykova, M.~Halinkovic, M.~Galinski, P.~Laskov, and A.~Vinel, ``Robots for safer pedestrian crossing on two-lane roads,'' in \emph{Proc. IEEE IAVVC}, 2025.

\bibitem{Alexey-ELMAR25}
{A. L. Flores Comeca, N. Masarykova, M. Halinkovic, M. Galinski, P. Laskov, and A. Vinel}, ``Social robots for road safety: Pedestrian crossing assistance use-case,'' in \emph{Proc. ELMAR}, 2025.

\bibitem{li2024bias}
X.~Li, Z.~Chen, J.~Zhang, F.~Sarro, Y.~Zhang, and X.~Liu, ``Bias behind the wheel: Fairness testing of autonomous driving systems,'' \emph{ACM Trans. Softw. Eng. Methodol.}, vol.~34, no.~3, pp. 1--24, 2025.

\bibitem{agebias}
M.~Brandao, ``Age and gender bias in pedestrian detection algorithms,'' \emph{arXiv preprint arXiv:1906.10490}, 2019.

\bibitem{augmentation}
C.~Chi, S.~Zhang, J.~Xing, Z.~Lei, S.~Li, and X.~Zou, ``Pedhunter: Occlusion robust pedestrian detector in crowded scenes,'' \emph{Proc. AAAI Conf. Artif. Intell.}, vol.~34, pp. 10\,639--10\,646, 2020.

\bibitem{fair-ped}
M.~Khoshkdahan, N.~Kjär, and F.~B. Flohr, ``Fair-ped: Fairness evaluation in pedestrian detection using clip,'' in \emph{\iv}, 2025, pp. 1504--1509.

\bibitem{braun2019eurocity}
M.~Braun, S.~Krebs, F.~Flohr, and D.~M. Gavrila, ``Eurocity persons: A novel benchmark for person detection in traffic scenes,'' \emph{\pami}, vol.~41, no.~8, pp. 1844--1861, 2019.

\bibitem{hardt2016equality}
M.~Hardt, E.~Price, and N.~Srebro, ``Equality of opportunity in supervised learning,'' in \emph{\nips}, 2016, pp. 1--9.

\bibitem{zhang2021ignorance}
J.~M. Zhang and M.~Harman, ``"ignorance and prejudice" in software fairness,'' in \emph{Proc. ICSE}, 2021, pp. 1436--1447.

\end{thebibliography}

\end{document}